\documentclass[letterpaper, 10 pt, conference]{ieeeconf}  

\IEEEoverridecommandlockouts                              

\usepackage{blindtext}
\usepackage{graphicx}
\usepackage{epstopdf}
\usepackage{float}
\usepackage{setspace}
\usepackage{amsmath}

\usepackage{mathrsfs}
\graphicspath{ {./images/} }
\usepackage{amssymb}
\usepackage[noadjust]{cite}
\usepackage{multirow}
\usepackage{array}
\usepackage{inputenc}
\let\labelindent\relax
\usepackage[inline]{enumitem}
\usepackage{soul}
\usepackage{booktabs}
\usepackage{varwidth}
\usepackage{bm}

\usepackage{hyperref}
\usepackage{tikz}

\usepackage{amsthm}

\newtheorem{assumption}{{Assumption}}

\newcommand{\bbm}{\begin{bmatrix}}
\newcommand{\ebm}{\end{bmatrix}}

\newcommand{\Real}{\mathbb{R}}

\newcommand{\lone}{{\mathcal{L}_1}}
\newcommand{\linf}{{\mathcal{L}_\infty}}

\newcommand{\lnone}[1]{\left\Vert#1\right\Vert_{\lone}}
\newcommand{\lninf}[1]{\left\Vert#1\right\Vert_{\linf}}

\setlist*[enumerate,1]{%
  label=(\roman*),
}

\title{\LARGE \bf
Adaptive Model Predictive Control for High-Accuracy Trajectory Tracking in Changing Conditions
}

\author{Karime Pereida and Angela P. Schoellig
\thanks{The authors are with the Dynamic Systems Lab (www.dynsyslab.org) at the University of Toronto Institute for Aerospace Studies (UTIAS), Canada. 
        Email: {k.pereidaperez@mail.uto\-ronto.ca, schoellig@utias.utoronto.ca} }%
\thanks{This research was supported in part by the Mexican National Council of Science and Technology (abbreviated CONACYT).}
}

\newcommand\copyrighttext{\footnotesize \textbf{Sub version.} Accepted at \textit{2018 IEEE/RSJ International Conference on Intelligent Robots and Systems.}

\textcopyright 2018 IEEE. Personal use of this material is permitted. Permission from IEEE must be obtained for all other uses, in any current or future media, including reprinting/republishing this material for advertising or promotional purposes, creating new collective works, for resale or redistribution to servers or lists, or reuse of any copyrighted component of this work in other works.}

\newcommand\copyrightnotice{\begin{tikzpicture}[remember picture,overlay]
\node[anchor=south,yshift=10pt] at (current page.south) {\fbox{\parbox{\dimexpr\textwidth-\fboxsep-\fboxrule\relax}{\copyrighttext}}};
\end{tikzpicture}}

\begin{document}

\maketitle
\copyrightnotice{} 
\thispagestyle{empty}
\pagestyle{empty}

\begin{abstract}
Robots and automated systems are increasingly being introduced to unknown and dynamic environments where they are required to handle disturbances, unmodeled dynamics, and parametric uncertainties. Robust and adaptive control strategies are required to achieve high performance in these dynamic environments. In this paper, we propose a novel adaptive model predictive controller that combines model predictive control (MPC) with an underlying $\lone$ adaptive controller to improve trajectory tracking of a system subject to unknown and changing disturbances. The $\lone$ adaptive controller forces the system to behave in a predefined way, as specified by a reference model. A higher-level model predictive controller then uses this reference model to calculate the optimal reference input based on a cost function, while taking into account input and state constraints. We focus on the experimental validation of the proposed approach and demonstrate its effectiveness in experiments on a quadrotor. We show that the proposed approach has a lower trajectory tracking error compared to non-predictive, adaptive approaches and a predictive, non-adaptive approach, even when external wind disturbances are applied. 

\end{abstract}

\section{INTRODUCTION}

Robots and automated systems are increasingly being introduced to unknown and dynamic environments, which requires the design of sophisticated control methods that can achieve high overall performance in these situations. Unlike traditional controllers where small changes in the conditions may significantly deteriorate the controller performance and may cause instability (see~\cite{Skelton1989},~\cite{Morari1999}, and~\cite{Skogestad2007}), controllers deployed in changing environments must be robust to model uncertainties, unknown disturbances, and changing dynamics. Robotic applications in these increasingly challenging environments include autonomous driving, assistive robotics, and unmanned aerial vehicle (UAV) applications. For example, in airborne package delivery, packages to be delivered can have different properties (mass, center of gravity, and inertia) and the vehicles may encounter different weather conditions. As such, it is not feasible to design a controller for each environmental or system condition. 

\begin{figure}[t]
   \centering
   \includegraphics[width=0.45\textwidth]{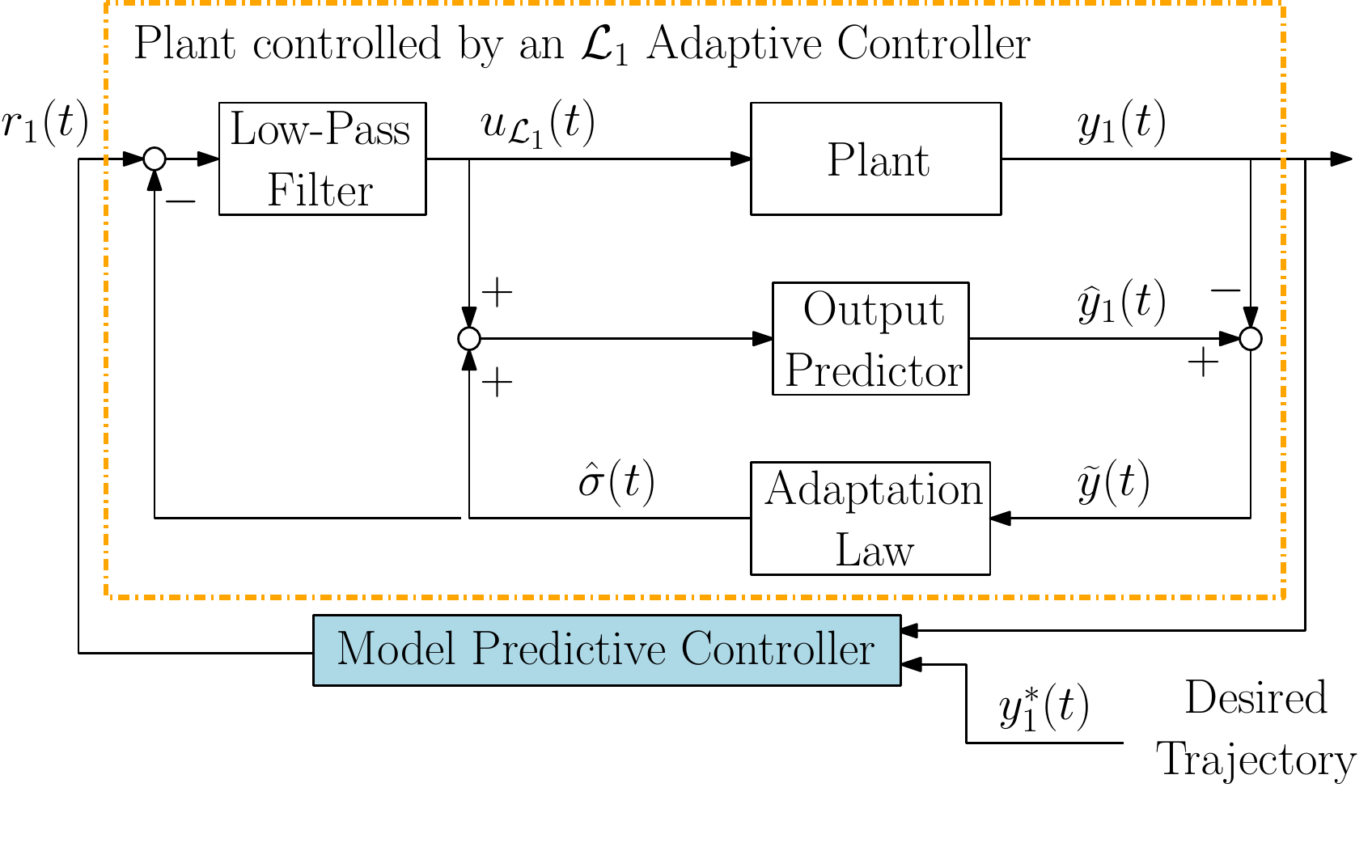}
   \vspace{-0.7cm}
   \caption{The proposed adaptive model predictive control architecture. The underlying $\lone$ adaptive controller forces the system to behave close to a specified linear reference model even in the presence of uncertainties and disturbances. Using the specified linear reference model, the model predictive controller calculates at each time step an input to the plant controlled by the $\lone$ adaptive controller (which now behaves as the specified reference model) such that the trajectory tracking error is minimized.}
   \vspace{-14pt}
   \label{fig:blockdiagram}
\end{figure}

In this paper, we present a controller that achieves high-accuracy tracking performance and is robust to unknown disturbances and changing dynamics. We consider nonlinear systems and propose a novel architecture that combines $\lone$ adaptive control and model predictive control (MPC), see Fig.~\ref{fig:blockdiagram}. MPC is an attractive control scheme because constraints can easily be incorporated and because of its predictive capability. It is an optimal control scheme; however, its performance depends on the accuracy of the model used in the optimization. To overcome this, we propose an underlying $\lone$ adaptive controller, which forces a potentially nonlinear system to behave in the same predefined way, as specified by a linear reference model, even in the presence of model uncertainties. As a result, a standard linear MPC is able to achieve high-accuracy trajectory tracking of the overall system. We validate the proposed approach through extensive experiments on a quadrotor. We show that the predictive capabilities of MPC and the robustness to disturbances of the $\lone$ adaptive control enable the proposed approach to achieve better trajectory tracking performance compared to non-predictive and/or non-adaptive approaches, especially when external, unknown disturbances are present.

Model predictive control solves an optimization problem at each time step, based on  a model of the system and the current state of the system, to find an optimal control sequence that minimizes the given objective function subject to input and state constraints. The first input value of that sequence is applied to the system. In the next time step, the optimization problem is solved again using the new state of the system. MPC has been widely applied in practice, where satisfaction of constraints is important~\cite{Mayne2000}. The standard implementation of MPC using a nominal model of the system dynamics exhibits nominal robustness to small disturbances~\cite{Marruedo2002}. However, these robustness guarantees may be insufficient in practical situations where larger disturbances may be present. 

To handle model uncertainties, adaptive MPC schemes have been introduced, which update the model online based on measurement data. In~\cite{Fukushima2007}, an adaptive MPC method for a class of constrained linear, time-invariant systems is proposed. This approach is based on a novel method to estimate the parameters of a model suitable for MPC. The algorithm is initially conservative due to large parameter uncertainties, but performance is improved over time as the parameter estimate’s uncertainty is reduced and the estimate converges to the true value. However, the quality of the estimation depends on the excitation of the state. Similar adaptive approaches have been proposed for nonlinear systems; however, these are challenging to design and computationally expensive. An adaptive nonlinear MPC scheme for constrained nonlinear systems is presented in~\cite{Adetola2009}. It uses min-max nonlinear MPC in combination with an adaptation mechanism to address the issue of robustness while the estimated value is evolving. Once again, the algorithm is initially conservative due to large parameter uncertainties, and the conservativeness reduces as the parameter estimate improves. However, this approach is still computationally expensive. Finally, learning-based MPC approaches have used neural networks (see~\cite{Wang2016} and~\cite{Lenz2015}) or Gaussian processes (see~\cite{Ostafew2016} and~\cite{Mckinnon2018}) to learn the dynamics of the system used in the controller. These approaches require a significant amount of data in order to build an accurate model and often do not adapt to changes in the environment in real time and at high update rates. 

Further, $\lone$ adaptive control is based on the model reference adaptive control (MRAC) architecture but includes a low-pass filter that decouples robustness from adaptation~\cite{Hovakimyan2010}. As a result, arbitrarily high adaptation gains can be chosen for fast adaptation. It has successfully been applied to fixed-wing vehicles~\cite{Beard2006},~\cite{Kaminer2007}, quadrotors~\cite{Zuo2014},~\cite{Pereida2017}, a NASA AirSTAR flight test vehicle~\cite{Gregory2009}, a tailless unstable aircraft~\cite{Patel2007}, and hexacopter and octocopter vehicles~\cite{Mallikarjunan2012}. The $\lone$ adaptive controller forces a system, which may be nonlinear and subject to model uncertainties, to behave as a predefined linear reference model. Leveraging this characteristic, it was used in combination with iterative learning control (ILC), where ILC enabled the system to improve trajectory tracking over iterations~\cite{Pereida2017}. ILC computes the input for the next iteration offline after the previous iteration is completed. In contrast to~\cite{Pereida2017}, in this work, we calculate the inputs online using MPC, which enables the system to achieve high-accuracy trajectory tracking on the first iteration.

{{The key contributions of this paper are:
\begin{itemize}
 \item to demonstrate the effectiveness of a combined $\lone$ adaptive control and a linear MPC approach to control nonlinear systems with reduced computational cost as compared to nonlinear MPC schemes; 
 \item to remove the need for persistent excitation to achieve accurate adaptation as in existing adaptive MPC strategies~\cite{Marafioti2014};
 \item to validate the novel $\lone$ adaptive control and MPC approach through extensive experimental results;
 \item to demonstrate the robustness of the proposed approach to external disturbances, (e.g. wind), as shown in experimental results with a quadrotor;
 \item to demonstrate the improved tracking performance on a quadrotor compared to two non-predictive and adaptive frameworks, and a predictive and non-adaptive scheme. 
\end{itemize} }}

The remainder of this paper is organized as follows. The problem is defined in Section~\ref{sec:Problem}. The details of the proposed approach are presented in Section~\ref{sec:Methodology}. Experimental results on a quadrotor are presented in Section~\ref{sec:Experimental}. Conclusions are provided in Section~\ref{sec:Conclusions}.

\section{PROBLEM STATEMENT}
\label{sec:Problem}
The objective of this work is to achieve high-accuracy trajectory tracking in the presence of uncertain, and possibly changing conditions on the first iteration. Consider a system whose dynamics (`Plant' block in Fig.~\ref{fig:blockdiagram}) are unknown, can change over time, and can be described by a single-input, single-output (SISO) system (this approach is later extended to multi-input, multi-output (MIMO) systems) identical to~\cite{Hovakimyan2010} for output feedback: 
\begin{equation}
 \begin{array}{c}
  y_1(s) = A(s)(u_\lone(s) + d_\lone(s))\,, 
 \end{array}
 \label{eq:Asystem}
\end{equation}
where $y_1(s)$ is the Laplace transform of the output $y_1(t)$, $A(s)$ is a strictly-proper unknown transfer function that can be stabilized by a proportional-integral controller, $u_\lone(s)$ is the Laplace transform of the input signal, and $d_\lone(s)$ is the Laplace transform of the disturbance signal defined as $d_\lone(t) \triangleq f(t,y_1(t))$, where $f : \Real \times \Real \rightarrow \Real $ is an unknown map subject to the following assumption:

\begin{assumption}[Lipschitz continuity]
\label{as:lipshitz}
There exist constants $L > 0$ and $L_0 > 0$, such that the following inequalities hold uniformly for all $t$:
\begin{align}
    | f(t,v) - f(t,w) | &\leq L | v - w | ~ \text{, and} \label{eq:assumption1} \\
    | f(t,w) | &\leq L | w | + L_0 ~. \label{eq:assumption2}
\end{align}
\end{assumption}

The system is required to accurately track a desired trajectory $y^*_1(t)$ defined over a finite time interval. Given $y^*_1(t)$, we aim to compute $u_{\lone}$ such that the system~\eqref{eq:Asystem} is able to achieve high-accuracy trajectory tracking at a low computational cost. To achieve this, we propose a control architecture 
as shown in Fig.~\ref{fig:blockdiagram}, where a model predictive controller is used on top of an underlying $\lone$ adaptive controller.

\section{METHODOLOGY}
\label{sec:Methodology}
We consider two main subsystems: the $\lone$ adaptive controller (orange dashed box in Fig.~\ref{fig:blockdiagram}) and the model predictive controller. The $\lone$ adaptive controller is presented in Section~\ref{ssec:L1AC}. Section~\ref{ssec:MPC} introduces the MPC. The combined $\lone$ adaptive control and MPC framework is described in Section~\ref{ssec:L1MPC}. 

\subsection{$\lone$ Adaptive Control}
\label{ssec:L1AC}
The $\lone$ adaptive controller is used to make the system behave as a predefined, linear system even when unknown, changing disturbances act on the system. In this work, the standard $\lone$ adaptive output feedback controller for SISO systems~\cite{Hovakimyan2010} is used. For convenience and completeness, we present a brief summary of the $\lone$ adaptive control scheme. 

\emph{Problem Formulation:} The output feedback $\lone$ adaptive controller computes a control input $u_\lone(t)$ such that $y_1(t)$ tracks the bounded piecewise continuous reference input $r_1(t)$ according to the following first-order reference system: 
\begin{equation}
 M(s) = \frac{m}{s+m}\,,\quad m>0\,.
 \label{eq:refsys}
\end{equation}

\emph{Definitions and $\lone$-Norm Condition:} We can rewrite system~\eqref{eq:Asystem} in terms of the reference system~\eqref{eq:refsys} as follows: 
\begin{equation}
 y_1(s) = M(s) (u_\lone(s) + \sigma(s))\,,
 \label{eq:Msystem}
\end{equation}
where uncertainties in $A(s)$ and $d_\lone(s)$ are combined into $\sigma(s)$: 
\begin{equation}
 \sigma(s) \triangleq \frac{(A(s)-M(s))u_\lone(s)+A(s)d_\lone(s)}{M(s)}\,.
\end{equation}
The low-pass filter $C(s)$ (see Fig.~\ref{fig:blockdiagram}) is assumed to be strictly proper with $C(0)=1$, with: 
\begin{equation}
 H(s) \triangleq \frac{A(s)M(s)}{C(s)A(s)+(1-C(s))M(s)}\quad \text{being stable,}
\end{equation}
and the following $\lone$-norm condition being satisfied: 
\begin{equation}
 \lnone{G(s)}L<1\,,\ \ \text{where } G(s)\triangleq H(s)(1-C(s))\,,
 \label{eq:l1norm}
\end{equation}
where $L$ is the Lipschitz constant defined in Assumption~\ref{as:lipshitz}.


The above assumption holds for cases when $A(s)$ can be stabilized by a proportional-integral controller, as shown in~\cite{Hovakimyan2010}. The equations that describe the implementation of the $\lone$ output feedback architecture are presented below.
\begin{description}
 \item [Output Predictor:] The output predictor is used to estimate the system output based on the reference model~\eqref{eq:Msystem}: 
 \[
  \dot{\hat{y}}_1(t) = -m\hat{y}_1(t)+m(u_\lone(t)+\hat{\sigma}(t))\,, \quad \hat{y}_1(0)=0\,,
 \]
where $\hat{\sigma}(t)$ is the estimate of $\sigma(t)$. In the Laplace domain,  
\begin{equation}
 \hat{y}_1(s) = M(s)(u_\lone(s)+\hat{\sigma}(s))\,.
 \label{eq:outpredict}
\end{equation}
\item [Adaptation Law:] The following adaptation law is used to update the estimate $\hat{\sigma}(t)$: 
\begin{equation}
 \dot{\hat{\sigma}}(t) = \Gamma\, \text{Proj} (\hat{\sigma}(t), -\tilde{y}(t))\,, \quad \hat{\sigma}(0)=0\,,
 \label{eq:adaptation}
\end{equation}
where $\tilde{y}(t)\triangleq \hat{y}_1(t)-y_1(t)$, $\Gamma \in \mathbb{R}^+$ is the adaptation rate subject to the lower bound specified in~\cite{Hovakimyan2010}. To ensure fast adaptation, the adaptation rate is chosen to be very large. The projection operator $\text{Proj}(\cdot)$, as defined in~\cite{Hovakimyan2010}, guarantees that the estimated value of $\sigma(t)$ remains within a specified convex set. 
\item [Control Law:] The control input is calculated by passing the difference between the desired trajectory signal $r_1(t)$ and the adaptive estimate $\hat{\sigma}(t)$ through a low-pass filter:
\begin{equation}
 u_\lone(s) = C(s)(r_1(s)-\hat{\sigma}(s))\,.
 \label{eq:input}
\end{equation}
This means that the $\lone$ adaptive controller compensates only for low-frequency uncertainty components of $A(s)$ and $d_\lone(s)$, i.e., the frequencies that the system is able to counteract. The low-pass filter attenuates high-frequency components.
\end{description}

\emph{Performance Bounds:} We choose the $\lone$ adaptive controller as it provides performance bounds on the estimation and output errors. The closed-loop reference system for the $\lone$ adaptive controller is described as follows: 
\begin{equation}
 y_{1,ref}(s) = M(s)(u_{ref}(s)+\sigma_{ref}(s))\,
 \label{eq:y1ref}
\end{equation}
\begin{equation}
 u_{ref}(s) = C(s)(r_1(s)-\sigma_{ref}(s))\,,
 \label{eq:uref}
\end{equation}
where 
\[
 \sigma_{ref}(s)= \frac{(A(s)-M(s))u_{ref}(s)+A(s)d_{ref}(s)}{M(s)}\,
\]
and $d_{ref}(s)$ is the Laplace transform of $d_{ref}(t)\triangleq f(t,y_{1,ref}(t))$. If $C(s)$ and $M(s)$ satisfy the $\lone$-norm condition~\eqref{eq:l1norm}, then the closed-loop reference system described in~\eqref{eq:y1ref} and~\eqref{eq:uref} is bounded-input, bounded-output (BIBO) stable.

In~\cite{Hovakimyan2010}, it is shown that when system~\eqref{eq:Asystem} is subject to the $\lone$ output feedback adaptive controller described in~\eqref{eq:outpredict},~\eqref{eq:adaptation}, and~\eqref{eq:input}, and $C(s)$ and $M(s)$ satisfy the $\lone$-norm condition in~\eqref{eq:l1norm}, the following bounds hold: 
\begin{equation}
 \lninf{\tilde{y}}<\gamma_0\,,
 \label{eq:gamma0}
\end{equation}
\begin{equation}
 \lninf{y_{1,ref}-y_1}<\gamma_1\,,
 \label{eq:gamma1}
\end{equation}
where $\tilde{y}(t)\triangleq \hat{y}_1(t)-y_1(t)$, $\gamma_0$ is defined in equation~(4.31) in~\cite{Hovakimyan2010} and 
\[
 \gamma_1 \triangleq \frac{\lnone{\frac{C(s)H(s)}{M(s)}}}{1-\lnone{G(s)}L}\gamma_0\,.
\]
It is important to note that $\gamma_0\propto\sqrt{\frac{1}{\Gamma}}$, which means that for high adaptation gains, the estimation error is bounded and can be made arbitrarily small. Furthermore, the actual system output approaches the behavior of the reference system~\eqref{eq:y1ref}. 

When the uncertainties are within the bandwidth of the low-pass filter, the controller is able to cancel the uncertainties exactly. In this ideal scenario, the system response is the following: 
\begin{equation}
 y_{1,id}(s) = M(s) r_1(s)\,.
 \label{eq:yid}
\end{equation}
Even when the ideal case is not achieved, the transient and steady-state performance of the controller have the performance guarantees provided in~\eqref{eq:gamma0} and~\eqref{eq:gamma1}. For the proposed $\lone$-MPC framework, we require that the system controlled by the $\lone$ adaptive controller behaves as close as possible to a known, linear model. As shown in~\eqref{eq:gamma0}, in order to achieve this, we must choose $\Gamma$ to be very large.

\emph{Multi-Input, Multi-Output Implementation}
The SISO architecture described above can be extended to a MIMO implementation. We assume that the states are decoupled, which can be achieved after applying an appropriate feedback linearization. Then, for $n$ different outputs, the low-pass filter $C(s)$ and the first-order reference system~\eqref{eq:refsys} are implemented as $(n\times n)$ diagonal transfer function matrices:
\begin{equation}
\def\arraystretch{1.4}
\begin{array}{rcl}
\label{eq:CM}
  C(s) & = & \text{diag}(C_1(s), \hdots, C_n(s))\,, \\
  M(s) & = & \text{diag}(M_1(s), \hdots, M_n(s))\,, 
\end{array}
\end{equation}
where $C_i(s) = \frac{\omega_i}{s+\omega_i}$, $\omega_i>0$, for example, $M_i(s)=\frac{m_i}{s+m_i}$, $m_i>0$, and $i=1,\hdots,n$. 

\subsection{Model Predictive Control}
\label{ssec:MPC}
MPC finds an optimal control input sequence that steers the system towards a desired trajectory, taking into account input and state constraints. The algorithm updates the control signal at every sampling time based on a known system model and desired future outputs. The key dynamics of the system to be controlled can be described by the following linear, discrete-time model:
\begin{equation}
\def\arraystretch{1.4}
\begin{array}{l}
y_1(k+1) = Ay_1(k) + Br_1(k)\,, \ y_1(0)=y_{1,0}\,, \\ 
\end{array}
\label{eq:systemMPC}
\end{equation}
where $y_1(k)\in \mathbb{R}^{n}$ is the output at time step $k$, and $r_1(k)\in\mathbb{R}^n$ is the control input. The desired output trajectory $y_1^*(k)$ is assumed to be feasible based on the model~\eqref{eq:systemMPC}, where $(r_1^*(k),y_1^*(k))$ satisfy~\eqref{eq:systemMPC}. At every time step $\bar{k}$, MPC solves the following open-loop optimization problem: 
\begin{equation}
 \min_{r_1(\bar{k}),\hdots,r_1(\bar{k}+N_h)}J(y_1,r_1)
 \label{eq:cost}
\end{equation}
subject to
\[
 \begin{array}{l}
 y_1(k+1)  =  Ay_1(k) + Br_1(k)\,,\ \forall k=\bar{k},\hdots,\bar{k}+N_h\,,\\
 y_1(\bar{k})=y_{1,\bar{k}}
 \end{array}
\]
where $J(y_1,r_1)$ is defined as: 
\begin{equation}
\def\arraystretch{1.6}
\begin{array}{l}
 J(y_1,r_1)  = \\
  \sum_{k=\bar{k}+1}^{k+N_h+1} (y_1^*(k)-y_1(k))^T Q (y_1^*(k)-y_1(k)) + \\
  \sum_{k=\bar{k}}^{k+N_h} r_1^T(k) R r_1(k) + \delta r_1^T(k)S\delta r_1(k)\,,
 \end{array}
\end{equation}
where $\delta r_1(k) = r_1(k)-r_1(k-1)$, $Q\,,R\,,S\in\mathbb{R}^{n\times n}$, and $N_h$ is the length of the prediction horizon. 

We apply the computed optimal $r_1(\bar k)$ to the system and re-solve the optimization problem at $k=\bar{k}+1$ using the obtained measurement $y_1(\bar{k}+1)$. 

\subsection{Adaptive Model Predictive Control}
\label{ssec:L1MPC}
The proposed architecture uses MPC with an underlying $\lone$ adaptive controller (see Fig.~\ref{fig:blockdiagram}), which guarantees that the controlled system behaves close to a reference model. The MPC updates the input signal to the underlying $\lone$ adaptive controller $r_1(k)$ at every sampling time based on a known system model and desired future outputs. We assume that the $\lone$ adaptive controller makes the system behave close to~\eqref{eq:yid}, which is achievable if the low-pass filter $C(s)$ in~\eqref{eq:CM} is chosen appropriately. In the MIMO case, the ideal system response is described by~\eqref{eq:yid} where $M(s)$ is a diagonal transfer function matrix defined as in~\eqref{eq:CM}. The ideal system response can then be discretized and written as: 
\begin{equation}
 y_1(k+1) = A_Dy_1(k)+B_D(r_1(k)-y_1(k))\,,
\end{equation} 
where $A_D$ and $B_D$ are the discretized state-space matrices related to $M(s)$. We can then reformulate the open-loop optimization problem as: 
\begin{equation} 
 \def\arraystretch{1.6} 
 \begin{array}{l}
  \min_{r_1(\bar{k}),\hdots,r_1(\bar{k}+N_h)}  \\
  \Big[\sum_{k=\bar{k}+1}^{\bar{k}+N_h+1} (y_1^*(k)-y_1(k))^T Q  (y_1^*(k)-y_1(k)) + \\
   \sum_{k=\bar{k}}^{\bar{k}+N_h} r_1^T(k) R r_1(k) + \delta r_1^T(k)S\delta r_1(k) \Big]
 \end{array}
 \label{eq:optimization}
\end{equation}
subject to 
\[
\def\arraystretch{1.4}
\begin{array}{rcl}
 y_1(k+1) & = & A_Dy_1(k)+B_D(r_1(k)-y_1(k))\,,\\
 y_1(\bar{k}) & = & y_{1,\bar{k}}\,, \\
 |\ddot{r}_1(k)| & \leq & r_{max}\,, \ \forall k=\bar{k},\hdots,\bar{k}+N_h\,,
 \end{array} 
\]
where the second constraint may be added to ensure a smooth input and $r_{max}$ is designed based on the system's physical constraints.

\section{EXPERIMENTAL RESULTS}
\label{sec:Experimental}
This section shows experimental results of the proposed framework combining $\lone$ adaptive control and MPC (MPC-$\lone$) applied to a quadrotor flying three-dimensional trajectories and experiencing dynamic disturbances. We assess two main aspects of the proposed framework:
\begin{enumerate*}[font=\itshape]
 \item predictive performance and 
 \item robustness to external disturbances, as compared to non-predictive and/or non-adaptive frameworks.
\end{enumerate*}

\subsection{Quadrotor Setup}

\begin{figure}[t]
   \centering
   \vspace{4pt}
   \includegraphics[width=0.45\textwidth]{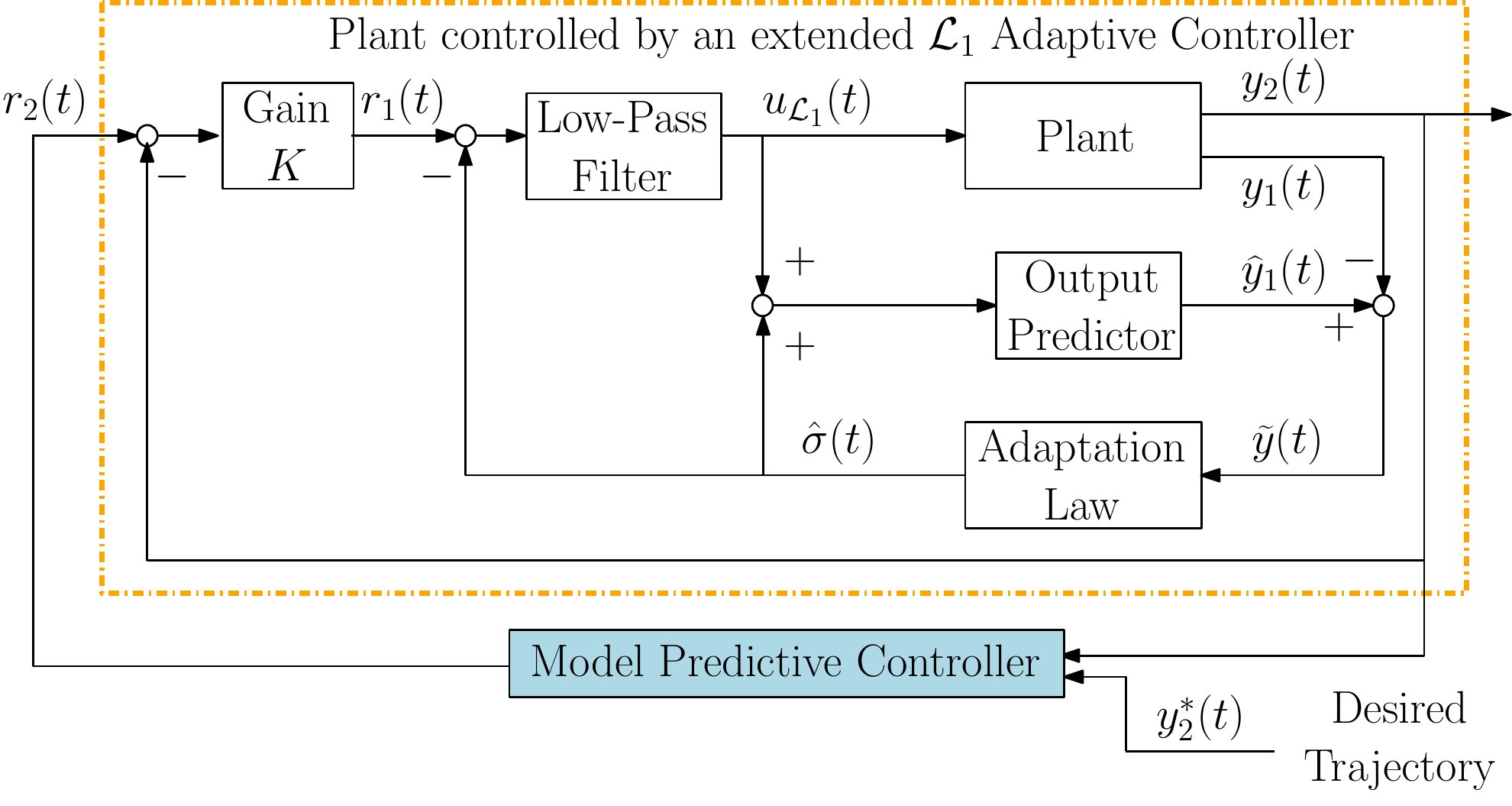}
   \caption{The proposed adaptive model predictive control architecture with an extended $\lone$ adaptive controller. The underlying $\lone$ adaptive controller has been augmented with an outer feedback loop with a proportional gain.}
   \vspace{-14pt}
   \label{fig:blockdiagramextended}
\end{figure}

The desired trajectory is described in the $x$, $y$ and $z$ directions. We implement the MIMO version of the $\lone$ adaptive controller described in Section~\ref{ssec:L1AC}, where we assume that the $x$, $y$ and $z$ directions are decoupled. We use a slightly modified $\lone$ adaptive controller structure (see Fig.~\ref{fig:blockdiagramextended}) as previously proposed in~\cite{Pereida2017},~\cite{Michini2009}. The difference is that this extended $\lone$ adaptive controller includes an additional outer feedback loop with a proportional gain. For the quadrotor example, we define $r_1(s)$ and $r_2(s)$ as the Laplace transforms of the desired translational velocity $r_1(t)\in\mathbb{R}^3$ and the desired bounded position reference input $r_2(t)\in\mathbb{R}^3$, respectively. We also  define $y_1(s)$ and $y_2(s)$ as the Laplace transforms of the vehichle's translational velocity $y_1(t)\in\mathbb{R}^3$ and position $y_2(t)\in\mathbb{R}^3$. 

We implement an outer-loop proportional controller as described in~\cite{Pereida2017},~\cite{Michini2009}, where
\[
 r_1(s) = K(r_2(s)-y_2(s))\,,
\]
where $K = \text{diag}(K_1,K_2,K_3)$ and $K_i>0$ for all $i=1,2,3$.
This extension to the $\lone$ adaptive controller results in the following ideal linear reference system: 
\begin{equation}
\def\arraystretch{1.4}
 y_{2,id}(s) = 
 \begin{bmatrix}
  D_1(s) & 0 & 0 \\
  0 & D_2(s) & 0 \\
  0 & 0 & D_3(s)
 \end{bmatrix}r_2(s)\,,
 \label{eq:sosystem}
\end{equation}
where
\begin{equation}
 D_i(s) = \frac{K_im_i}{s^2+m_is+K_im_i} \  \text{for } i={1,2,3}\,.\\
 \end{equation}
The system in~\eqref{eq:sosystem} is used to obtain $A_D$ and $B_D$ used for the equality constraints in the MPC implementation~\eqref{eq:optimization}. We use the IBM CPLEX solver to compute the desired positions $r_2(k)=(r_{2,x}(k),r_{2,y}(k),r_{2,z}(k))$. The interface to the real quadrotor requires commanded roll $\phi_{des}$, pitch $\theta_{des}$, vertical velocity $\dot{z}_{des}$, and yaw rate $\dot{\psi}_{des}$. Therefore, the signal $u_{\lone}(k)$ (see Fig.~\ref{fig:blockdiagramextended}) is transformed through the following equations: 
\[
 \begin{array}{l}
 \phi_{des} = - \text{arcsin}(-u_{\lone,x}\text{sin}(\psi)+u_{\lone,y}\text{cos}(\psi))\\
 \theta_{des} = \text{arcsin}(u_{\lone,x}\text{cos}(\psi)+u_{\lone,y}\text{sin}(\psi))\\
 \dot{z}_{des} = u_{\lone,z}\,,
 \end{array}
\]
where $\psi$ is the current yaw angle. The desired yaw angle $\psi_{des}$ is set to zero and controlled through a separate proportional controller.

\subsection{Experimental Setup}
\label{ssec:experimentalsetup}
\begin{figure}[t]
   \centering
   \includegraphics[width=0.45\textwidth]{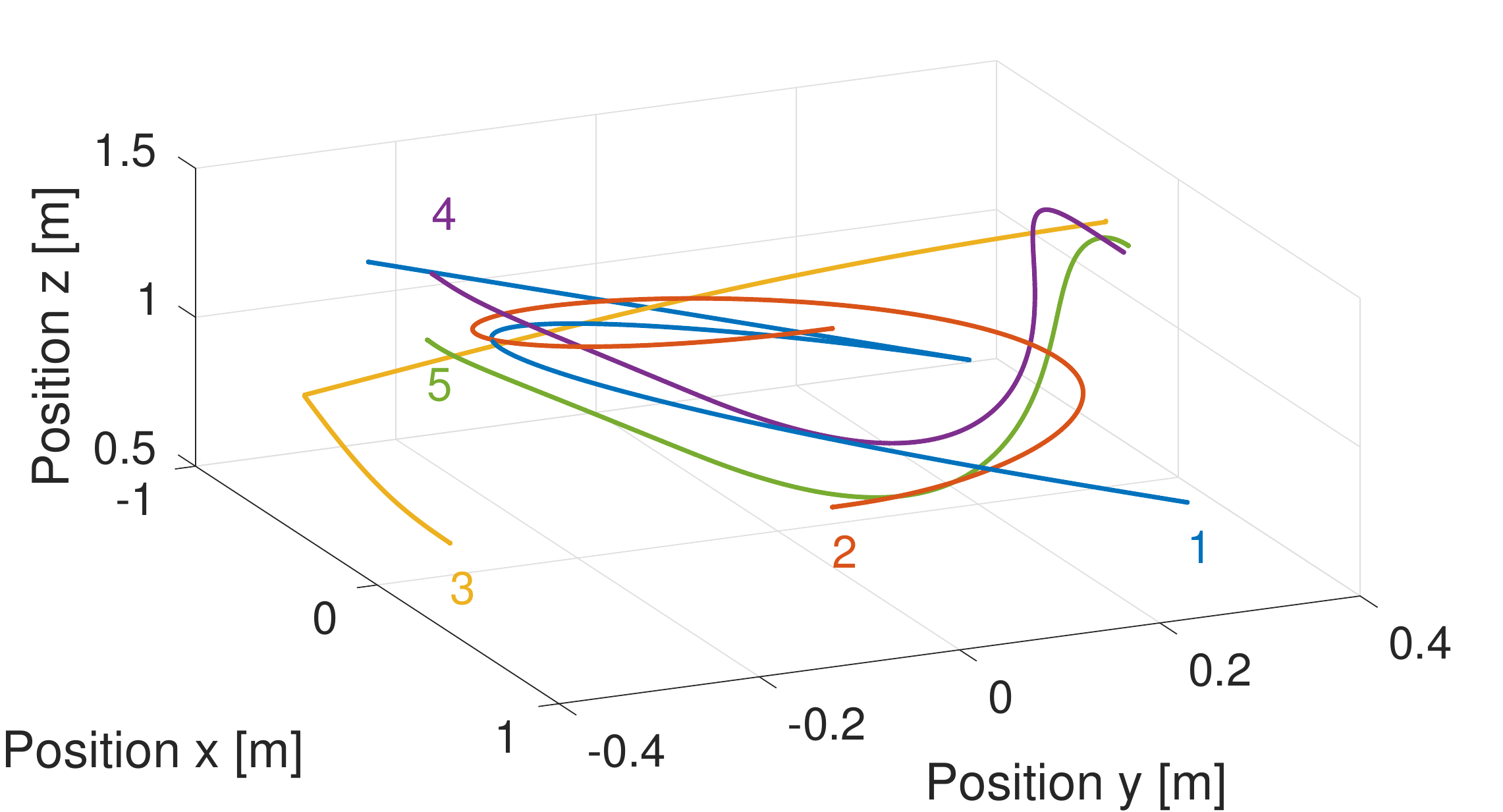}
   \caption{The five different trajectories that are used to test the MPC-$\lone$ control framework.}
   \vspace{-1pt}
   \label{fig:trajectories}
\end{figure}

The experiments were performed using the Parrot Bebop~2 quadrotor. An overhead motion capture camera system was used to obtain position, velocity, rotation, and rotational velocity measurements. Let $r_{2,j}(k)$ be the desired translational positions and $y_{2,j}(k)$ be the measured quadrotor positions in the $j=x,y,z$ directions, respectively. We propose five different trajectories to test our approach, as shown in Fig.~\ref{fig:trajectories}. To quantify the controller performance, we define the average position error as: 
\begin{equation}
 e = \frac{1}{N}\sum_{k=1}^{N}\sqrt{e_x^2(k)+e_y^2(k)+e_z^2(k)}\,,
\end{equation}
where $e_j(k) = r_{2,j}(k)-y_{2,j}(k)$, $N$ represents the trajectory length, and $j=x,y,z$.

The matrices of the cost function in~\eqref{eq:optimization} are defined as $Q=qI$, $R=rI$ and $S=sI$ where $q=17$, $r=0.08$ and $s=0.02$, and $I$ is the identity matrix. Since the objective is to improve the trajectory tracking performance, a significantly larger weight is given to this part in the cost function.

\subsection{$\lone$ Adaptive Control Performance Bounds}
\begin{figure}[t]
   \centering
   \includegraphics[width=0.5\textwidth]{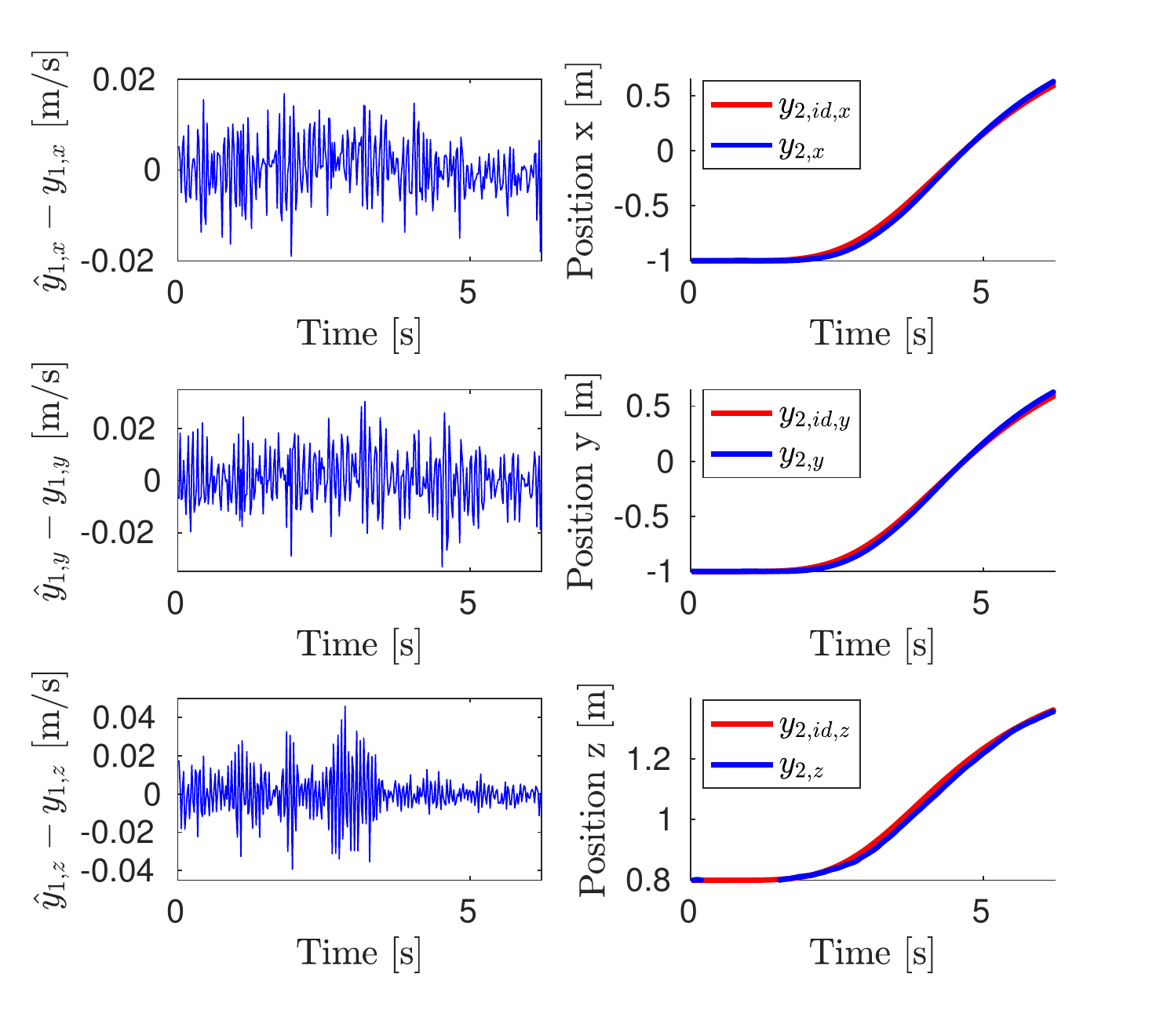}
   \vspace{-25pt}
  \caption{On the left, the estimation error $\hat{y}_{1,i}-y_{1,i}$ for each axis $i=x,y,z$ is shown. The estimation error remains bounded for the length of the trajectory, as suggested by~\eqref{eq:gamma0}. On the right, the ideal trajectory followed by the ideal system ~\eqref{eq:sosystem} (shown in red) is compared to the actual trajectory followed by the quadrotor (shown in blue).}
   \vspace{-10pt}
   \label{fig:yhaty2}
\end{figure}

In this subsection, we experimentally verify that 
\begin{enumerate*}[font=\itshape]
 \item the estimation error $\tilde{y}$ remains bounded, and 
 \item the actual system remains close to the ideal linear system for the extended $\lone$ adaptive controller described in~\cite{Pereida2017} and shown in Fig.~\ref{fig:blockdiagramextended}.
\end{enumerate*}
To do this, we use the reference input $r_2$ to command the quadrotor to hover for 1.5 seconds after which it starts moving in a 3D straight line. Fig.~\ref{fig:yhaty2} shows, on the left side, the estimation error $\hat{y}_{1,i}-y_{1,i}$ for each axis $i=x,y,z$. It is clear that for all three axes the estimation error remains bounded. Moreover, using the ideal linear reference model~\eqref{eq:sosystem}, we calculate the trajectory $y_{2,id}$ that the ideal reference system would follow if the reference input $r_2$ was applied. The right side of Fig.~\ref{fig:yhaty2} shows in red the trajectory of the ideal reference system $y_{2,id}$ and in blue the actual trajectory that the quadrotor followed $y_2$, when the input $r_2$ is applied. The actual system does behave close to the ideal reference system. 
  
\begin{figure}[t]
   \centering
   \includegraphics[width=0.45\textwidth]{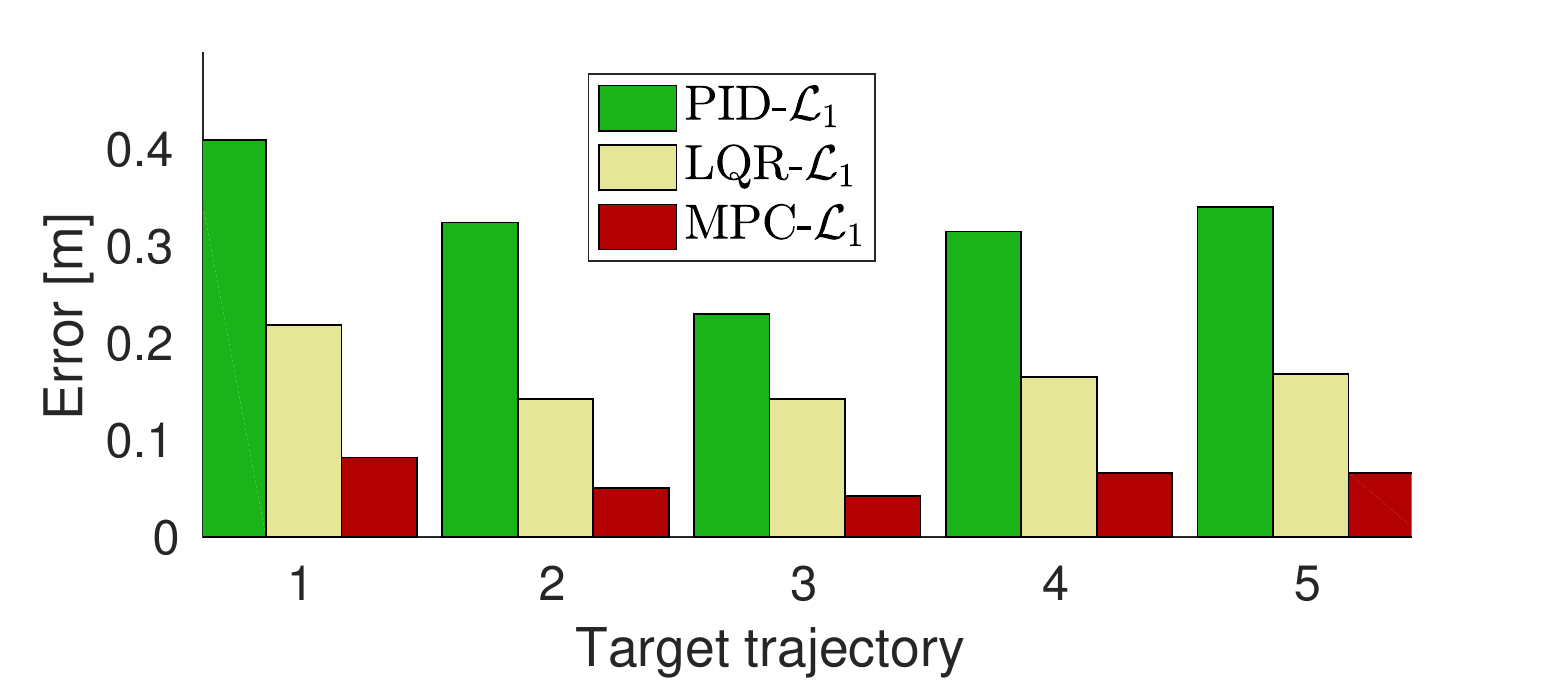}
   \caption{Average position tracking errors over five different trajectories for each of the three controllers: PID-$\lone$, LQR-$\lone$ and MPC-$\lone$. The PID- and LQR-based controllers use feedback information only; hence, the tracking errors are larger. The predictive controller is able to optimize the reference input based on future information of the desired trajectory. Consequently, the error achieved is significantly smaller.}
   \vspace{-1pt}
   \label{fig:pid_lqr_mpc_l1}
\end{figure}

\begin{figure*}[t]
   \centering
   \includegraphics[width=0.95\textwidth]{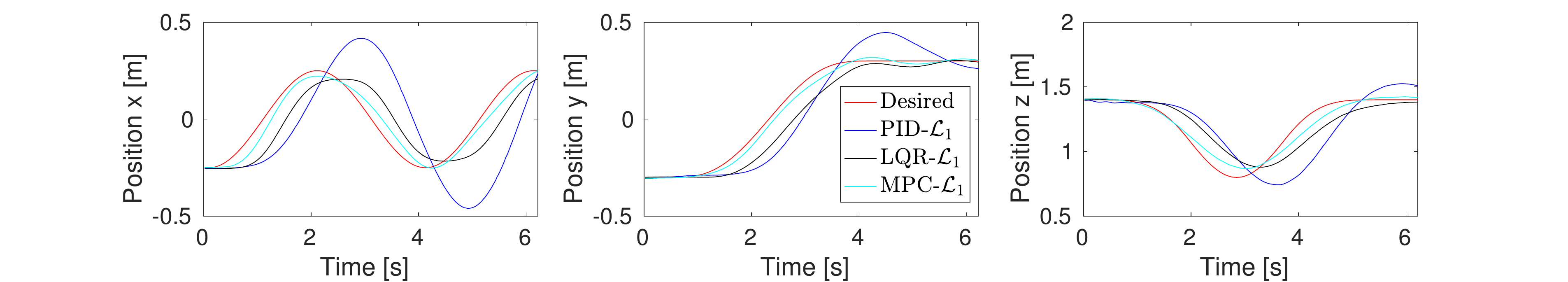}
   \caption{Actual position of the quadrotor tracking the purple trajectory in Fig.~\ref{fig:trajectories} for each of the three controllers: PID-$\lone$, LQR-$\lone$ and MPC-$\lone$. The feedback-only controllers PID-$\lone$ and LQR-$\lone$ present tracking delay. The proposed MPC-$\lone$ framework reduces the tracking error as it is able to include future desired outputs in the calculation of the optimal input.}
   \vspace{-1pt}
   \label{fig:timeresponse}
\end{figure*}

\begin{figure}[t]
   \centering
   \includegraphics[width=0.5\textwidth]{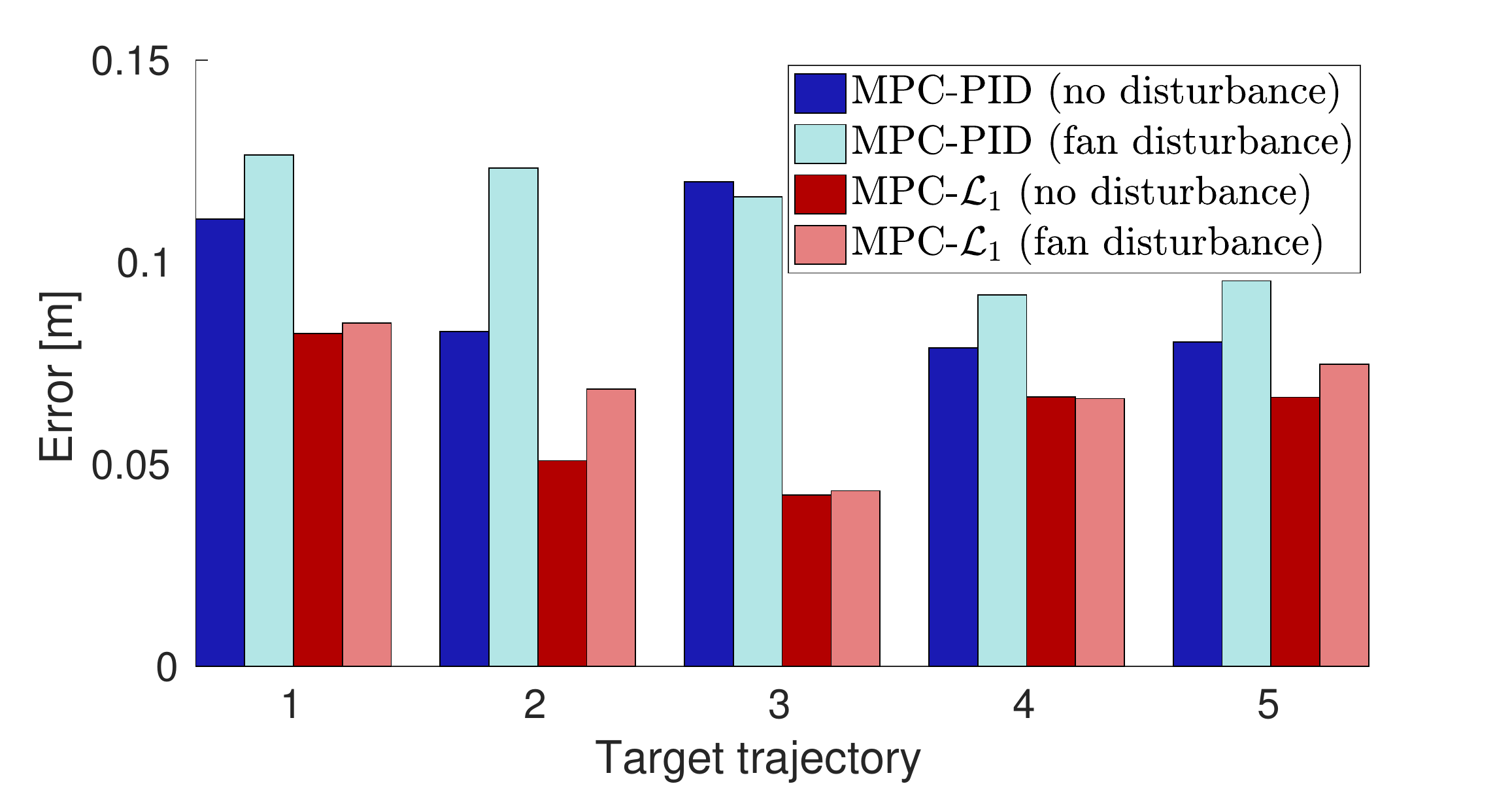}
   \caption{Average position tracking errors over five different trajectories when using the  MPC-PID and MPC-$\lone$ controllers. The cases when no disturbance is applied to the system are depicted in dark blue (MPC-PID) and dark red (MPC-$\lone$). A fan is used to apply a wind disturbance to the system and the resulting errors are depicted in light blue (MPC-PID) and light red (MPC-$\lone$). The proposed MPC-$\lone$ framework is able to compensate for the disturbance applied and has a performance similar to the case when no disturbance is applied. }
   \vspace{-1pt}
   \label{fig:mpc_pid_l1}
\end{figure}

\subsection{Predictive Performance}
We first demonstrate the benefit of the predictive controller component. To do this, we compare the tracking performance obtained with the proposed MPC-$\lone$ approach to that obtained with a PID-$\lone$ and LQR-$\lone$ approach. The PID-$\lone$ framework uses a proportional-integral-derivative (PID) controller to modify the input $r_2$ of the underlying $\lone$ adaptive controller. The LQR-$\lone$ framework uses an infinite-horizon linear-quadratic regulator (LQR) with integral component to modify the input $r_2$. The integral component in the LQR is used to decrease the steady-state error. 
All three approaches MPC-$\lone$, PID-$\lone$ and LQR-$\lone$ use the same underlying extended $\lone$ adaptive controller. The MPC-$\lone$ uses the cost function specified in Section~\ref{ssec:experimentalsetup}. The outer PID and LQR controllers were tuned in experiment to maximize the tracking performance for trajectory 4 (purple) in Fig.~\ref{fig:trajectories}. It is important to note that neither the PID-$\lone$ nor the LQR-$\lone$ controllers take into account future desired outputs when computing the input to the underlying $\lone$ adaptive controller. 

We compare the three controllers on the five test trajectories shown in Fig.~\ref{fig:trajectories}. The average position tracking errors for each controller and trajectory are shown in Fig.~\ref{fig:pid_lqr_mpc_l1}. The PID- and LQR-based controllers use only feedback information to correct for tracking errors and have significantly larger tracking errors than the proposed MPC-$\lone$ framework. This is better observed in Fig.~\ref{fig:timeresponse}, where the positions followed by the quadrotor for each of the three controllers, PID-$\lone$, LQR-$\lone$ and MPC-$\lone$, when tracking trajectory 4 (purple) in Fig.~\ref{fig:trajectories} are shown. The feedback-only controllers show a significant tracking delay, see the dark blue and black lines. The predictive component of the proposed MPC-$\lone$ framework optimizes over the prediction horizon and achieves a better performance, as shown by the cyan line in Fig.~\ref{fig:timeresponse}.  This shows the benefit of the predictive MPC component of the proposed approach. 
\addtolength{\textheight}{-4.5cm} 

\subsection{Robustness to Disturbances}
To assess the robustness to disturbances, the proposed framework is compared to an MPC-PID framework. The MPC-PID framework uses an MPC to modify the input to an underlying PID controller. The system model used in the MPC is obtained by applying a step input to the $x$, $y$, and $z$ directions separately and characterizing the system response when the quadrotor is controlled by the PID controller. The system is assumed to be a second-order, linear system in each direction. We first use the MPC-PID and MPC-$\lone$ frameworks to track each of the five desired trajectories. The cost functions used in the MPC-PID and MPC-$\lone$ frameworks are the same. 

The average position errors for the MPC-PID and MPC-$\lone$ approaches are shown in Fig.~\ref{fig:mpc_pid_l1} in dark blue and dark red, respectively. The system model obtained through the step response experiments for the MPC-PID framework is not accurate enough to describe the system behavior well. We found that tuning this system model for each trajectory could improve performance for that trajectory; however, doing so affected the performance of the other trajectories. In order to fairly compare both approaches, the model used in the MPC-PID framework is kept constant across trajectories. 

Next, using a fan we introduce a wind disturbance at different points in each trajectory. The resulting average errors when the fan disturbance is applied are shown in Fig.~\ref{fig:mpc_pid_l1} in light blue and light red for the MPC-PID and MPC-$\lone$ frameworks, respectively. The proposed MPC-$\lone$ framework is able to keep approximately the same performance when a disturbance is applied since the underlying $\lone$ adaptive controller is able to compensate for it. The MPC-PID approach performs generally worse when wind is applied compared to without wind. Only for trajectory 3, the disturbance applied to the system steers the system in the direction that reduces the tracking trajectory error. Nevertheless, the proposed framework has, on average, a significantly smaller increase in the error when the disturbance is applied.

\section{CONCLUSIONS}
\label{sec:Conclusions}
In this paper, we introduce a novel adaptive MPC framework to improve trajectory tracking performance in the presence of disturbances and partially unknown dynamics. The framework relies on an $\lone$ adaptive controller to keep the system close to a predefined linear system behavior, despite the presence of disturbances and changing dynamics. The MPC is used to calculate an optimal reference input to the underlying $\lone$ adaptive controller based on the reference model from the $\lone$ adaptive controller. Finally, we validated the proposed approach through experiments on a quadrotor. In experiments, the proposed method improves the tracking performance of the system as compared to non-predictive approaches (PID-$\lone$ and LQR-$\lone$) and a non-adaptive approach (MPC-PID), even in the presence of strong wind disturbances. The proposed approach is able to effectively control nonlinear systems with a computationally-efficient, adaptive, linear MPC approach.



%


\bibliographystyle{IEEEtran}
\bibliography{IEEEabrv,root}


\end{document}